%% file: main.tex
\RequirePackage{amsmath}
\documentclass[runningheads]{llncs}
\usepackage[T1]{fontenc}
\usepackage{import}
\usepackage{tikz}
\usepackage{graphicx}
\usepackage[caption=false]{subfig}
\usepackage{hyperref}
\usepackage{color}
\usepackage{amssymb}
\usepackage{booktabs}
\usepackage{multirow}
\usepackage{bm}
\usepackage[normalem]{ulem}
\usepackage{xcolor}
\usepackage[linesnumbered,ruled,vlined]{algorithm2e}

\SetCommentSty{mycommfont}

\newcommand{\ie}{\textit{i}.\textit{e}.}
\newcommand{\eg}{\textit{e}.\textit{g}.}
\newcommand{\ours}{SADM}

    \makeatletter
\def\@fnsymbol#1{\ensuremath{\ifcase#1\or \dagger\or \ddagger\or
   \mathsection\or \mathparagraph\or \|\or **\or \dagger\dagger
   \or \ddagger\ddagger \else\@ctrerr\fi}}
    \makeatother
\newcommand{\x}{\mathbf{x}}
\newcommand{\xx}{\mathbf{X}}
\newcommand{\cc}{\mathbf{c}}
\newcommand{\C}{\mathcal{C}}
\newcommand{\M}{\mathcal{M}}
\newcommand{\N}{\mathcal{N}}
\newcommand{\F}{\mathcal{F}}
\newcommand{\xM}{\xx^{\M}}
\newcommand{\xC}{\xx^{\C}}
\newcommand{\xF}{\xx^{\F}}
\newcommand{\R}{\mathbb{R}}
\newcommand{\z}{\mathbf{z}}

\newcommand{\h}{\mathbf{h}}
\newcommand{\eps}{\boldsymbol{\epsilon}}
\begin{document}

\title{\ours: Sequence-Aware Diffusion Model for Longitudinal Medical Image Generation}
\titlerunning{\ours: Sequence-Aware Diffusion Model}

\author{
Jee Seok Yoon\inst{1,3}\orcidID{0000-0003-0721-504X} \and
Chenghao Zhang\inst{2}\orcidID{0000-0002-5252-0046} \and
Heung-Il Suk\inst{1}\orcidID{0000-0001-7019-8962}\and
Jia Guo\inst{2}\orcidID{0000-0003-4803-0279}\and
Xiaoxiao~Li\thanks{Corresponding Author}\inst{,3}\orcidID{0000-0002-8833-0244}
}
% \author{
% Jee Seok Yoon\inst{1,4} \and
% Chenghao Zhang\inst{3} \and
% Heung-Il Suk\inst{1,2}\and
% Jia Guo\inst{3}\and
% Xiaoxiao~Li\inst{4}
% }
% \author{Jee Seok Yoon\inst{1,4}\orcidID{0000-0003-0721-504X} \and
% Chenghao Zhang\inst{3}\orcidID{} \and
% Heung-Il Suk\inst{1,2}\orcidID{0000-0001-7019-8962} \and
% Jia Guo\inst{3}\orcidID{0000-0003-4803-0279} \and
% Xiaoxiao Li\inst{4}\orcidID{0000-0002-8833-0244}}
%
\authorrunning{J. Yoon et al.}
% First names are abbreviated in the running head.
% If there are more than two authors, 'et al.' is used.
%

\institute{
Korea University, Seoul 02841 Republic of Korea\\
\email{\{wltjr1007,hisuk\}@korea.ac.kr} \and
Columbia University, NY 10027 USA\\
\email{\{cz2639,jg3400\}@columbia.edu}\and
The University of British Columbia, BC V6T 1Z4 Canada\\
\email{xiaoxiao.li@ece.ubc.ca}
}
% \institute{
% Department of Brain and Cognitive Engineering, Korea University, Seoul 02841, South Korea
% \email{\{wltjr1007,hisuk\}@korea.ac.kr} \and
% Department of Artificial Intelligence, Korea University, Seoul 02841, South Korea
% \and
% Department of Biomedical Engineering,  Columbia University, NY 10027 USA
% \email{\{cz2639,jg3400\}@columbia.edu}\and
% Department of Electrical and Computer Engineering, The University of British Columbia, BC V6T 1Z4 Canada
% \email{xiaoxiao.li@ece.ubc.ca}
% }

\maketitle

\input{section/0_abstract.tex}
\input{section/1_introduction.tex}
\input{section/2_background.tex}
\input{section/3_method.tex}
\input{section/4_experiment.tex}
\input{section/5_conclusion.tex}
\input{section/6_acknowledgement.tex}

\bibliographystyle{splncs04}
\bibliography{references}

\end{document}

%% file: section/0_abstract.tex
\begin{abstract}
Human organs constantly undergo anatomical changes due to a complex mix of short-term (\eg, heartbeat) and long-term (\eg, aging) factors.
Evidently, prior knowledge of these factors will be beneficial when modeling their future state, \ie, via image generation. 
However, most of the medical image generation tasks only rely on the input from a single image, thus ignoring the sequential dependency even when longitudinal data is available.
Sequence-aware deep generative models, where model input is a sequence of ordered and timestamped images, are still underexplored in the medical imaging domain that is featured by several unique challenges:
%. Unlike video generation, longitudinal medical image generation faces several unique challenges: 
1) Sequences with various lengths; 2) Missing data or frame, and 3) High dimensionality.
To this end, we propose a sequence-aware diffusion model (\ours) for the generation of longitudinal medical images. Recently, diffusion models have shown promising results in high-fidelity image generation. Our method extends this new technique by introducing a sequence-aware transformer as the conditional module in a diffusion model. The novel design enables learning longitudinal dependency even with missing data during training and allows autoregressive generation of a sequence of images during inference. 
% During training, our sequence-aware transformer learns to estimate the attentive representations in the longitudinal positions of given tokens based on the sequential input even with missing data. \XL{shorten}
% % Notably, our method can learn from sequences with missing data too.
% In inference time, we use an autoregressive sampling scheme to effectively generate new images. 
Our extensive experiments on 3D longitudinal medical images demonstrate the effectiveness of SADM compared with baselines and alternative methods. The code is available at \href{https://github.com/ubc-tea/SADM-Longitudinal-Medical-Image-Generation}{https://github.com/ubc-tea/SADM-Longitudinal-Medical-Image-Generation}. %

% \todo{
% I have addressed the necessary review comments of R1 and R4. Here are the ones remaining:
% \begin{itemize}
%     \item Adding code to the github link.
%     \item Table 2 does not say which dataset these numbers come from.
%     \item The review comments of R5.
% \end{itemize}
% }
%use the long-distance dependency between the source and generated images.
%To validate our work, we compare and ablate our proposed model over a toy face dataset and two medical image datasets, \ie, synthetic longitudinal brain MRI and ACDC cardiac MRI.

\keywords{Diffusion model \and Sequential image generation \and Autoregressive conditioning.}
\end{abstract}

%% file: section/1_introduction.tex
\section{Introduction}

% \todo{Fix intro
% 1. temporal/4d -> sequence
% 2. vs. video datasets
% 3. contribution
% 4. sequence-aware realted works in other domains, mention that not much work in generative models
% 5. long-term correspondence
% }

% Start: Why study image synthesis and progression?
% 1. Progression modeling -> 
% Examples: brain - age prograssion, heart - anatomical changes
% Why is this problem important?? Early diagnosis and cost of 
% 2. Image synthesis -> 

% Progression modeling is one of the most challenging tasks in medical image analysis mainly because the non-linear nature of age and disease affects people differently~\cite{Cole_2018}.
% It is also one of the most important procedures in prognosis and clinical planning as early treatment of disease (\eg, Alzheimer's~\cite{Rasmussen_2019}) can generally reduce its long-term severity.
% Progression modeling is one of the most important procedures in prognosis and clinical planning as it provides insight into human aging. However, it is also one of the most challenging tasks in the field mainly because the non-linear nature of age affects people and disease differently~\cite{Cole_2018}. 
% Progression modeling for medical image further requires predicting the anatomical changes of an organ after progression as short as a heartbeat to as long as lifetime of aging.
Iconic advancements of generative models in the medical domain have been possible due to several factors, such as state-of-the-art computational hardware and, more importantly, the availability of medical datasets (both open and in-house).
Hence, promising solutions for medical image synthesis, restoration, acceleration, and many other tasks have been proposed over the past few years~\cite{Yi_2019}.
Recent efforts to generate longitudinal medical images were mainly proposed for the following two tasks:
1) generation of longitudinal brain image~\cite{9854196}, which takes a source brain image and generates a new image with respect to chronological age (\ie, normal) progression or disease (\ie, abnormal) progression~\cite{causal_ml};
and 2) generation of multi-frame cardiac image~\cite{gan-based,diffusemorph}, which typically take a starting frame of a cardiac cycle (\ie, end-diastolic or ED phase) and generates the final frame of the cycle (\ie, end-systolic or ES phase).
An illustration showing examples of these tasks is presented in Fig.~\ref{fig:intro}.
Generative adversarial networks (GANs) have been a \textit{de facto} standard for these tasks in the past few years, but recent advances in diffusion models have shown promising results.
For example, the latent diffusion model, which uses the latent embedding of an image as input to the diffusion model to improve computational efficiency, has been used to synthesize high-quality 3D brain MRI~\cite{pinaya_2022}.
Similarly, a diffusion model was combined with a deformation image registration model, namely VoxelMorph~\cite{8633930}, to synthesize the end-systolic frame of cardiac MRI~\cite{diffusemorph,kim2022diffusion}.
However, most of these works rely on the input from a single image to generate its longitudinal images. Even when longitudinal samples are available, these methods often ignore the sequential dependency in the medical domain.
\input{figure/fig_intro.tex}

%Previous works on generative models for longitudinal medical image generation can largely be categorized into the following two branches: 1) Class-conditional generative models~\cite{gan-based,9854196}, which use a class (\eg, chronological age) or its embedding to control the generative process;
% and 2) Image-to-image (I2I) translation models~\cite{diffusemorph}, which take a source image and translate it to a target progression point (\eg, last frame of a cardiac cycle).

%Understandably, such longitudinal data has not been readily available in the past, and even the ones available today are scarce and have issues such as missing data and low temporal resolution.
% Unfortunately, past works proposed for these two branches only rely on a single image input without regard for temporal dependency.
Sequence-aware~\cite{Quadrana_2019} deep generative models are a class of generative models that can learn the sequential or temporal dependency of the longitudinal input data.
A sequence is defined as an ordered and timestamped set~\cite{Quadrana_2019}, and sequence-aware generative models take input from a sequence of images and output a generated image (formal definition in \textit{Problem settings} in Section~\ref{sec:method}).
Although such generative models for video datasets have been studied that often take a sequence of frames as input and learn their temporal dependence~\cite{vivit,imagen_video}, the existing solutions are not feasible for longitudinal medical data generation tasks. Because video datasets rarely have issues very common in the medical domain, such as \textit{1) longitudinal data scarcity, 2) missing frames or data, 3) high dimensionality, and 4) low temporal resolution}.

To this end, we explore ways to address these issues and propose a novel generative model for longitudinal medical image generation that can learn the temporal dependency given a sequence of medical images. 
% We have illustrated the difference between traditional generative models (\ie, class-conditional and I2I models), and sequence-aware generative models in Fig.~\ref{fig:compare_methods}\todo{New figure or not?}.
%To the best of our knowledge, we are the first to explore sequence-aware generative models for the medical image generation task and 
%To this end, we proposed a novel 
Our proposed method is named sequence-aware diffusion model (SADM). Specifically, during training, SADM learns to estimate attentive representations in the longitudinal positions of given tokens based on sequential input, even with missing data.
% Notably, our method can learn from sequences with missing data too.
In inference time, we use an autoregressive sampling scheme to effectively generate new images. Our extensive experiments on longitudinal 3D medical images demonstrate the effectiveness of SADM compared to baselines and alternative methods.
% Recent efforts to generate longitudinal medical image were mainly proposed for the following two tasks:
% 1) generation of longitudinal brain image~\cite{9854196}, which take a source brain image and generate a new image with respect to chronological age (\ie, normal) progression or disease (\ie, abnormal) progression~\cite{causal_ml};
% and 2) generation of multi-frame cardiac image~\cite{gan-based,diffusemorph}, which typically takes a starting frame of a cardiac cycle (\ie, end-diastolic phase) and generate the final frame of the cycle (\ie, end-systolic phase).
% Generative adversarial networks (GANs) have been a \textit{de facto} standard for these tasks in the past few years, but recent advances in diffusion models have shown promising results.
% For example, latent diffusion model, which uses the latent embedding of an image as input to the diffusion model to improve computational efficiency, has been used to synthesize high quality 3D brain MRI~\cite{pinaya_2022}.
% Similarly, a diffusion model was combined with a deformation image registration model, namely VoxelMorph~\cite{8633930}, to synthesize end-systolic frame of cardiac MRI~\cite{diffusemorph}.
%To this end, we propose the sequence-aware diffusion model (SADM) for longitudinal medical image generation.
The contributions of our SADM are as follows:

\begin{enumerate}
    \item To the best of our knowledge, we are one of the first to explore the temporal dependency of sequential data and use it as a prior in diffusion models for medical image generation.
    \item 
    %We propose novel sequence-aware diffusion model (SADM) for longitudinal medical image generation. 
    Our proposed SADM can work in various real-world settings, such as single image input, longitudinal data with missing frames, and high-dimensional images via the essential transformer module design.  
    \item We present state-of-the-art results in longitudinal image generation and missing data imputation for a multi-frame cardiac MRI and a longitudinal brain MRI dataset. 
\end{enumerate}

% Unfortunately, class-conditional methods are \sout{flawed as they rely on} chronological age which do not correspond well to the biological age progression~\cite{PH}.
% I2I models are limited in application as it can only translate images to a predefined progression point. 

% \XL{Do we want to focus on 4D or is 4D really important? Did you implement 3D or 2.5D on the cardiac data? }

% \XL{I suggest referring to the writing in this paper \href{https://openreview.net/pdf?id=0RTJcuvHtIu}{https://openreview.net/pdf?id=0RTJcuvHtIu}.}

% Fundamentally, it is desirable to have a spatially and temporally aware model for 4D medical image generation.
% However, most works assume the age progression is linear, and choose to interpolate between latent codes to generate intermediate temporal images.
% For example, DiffuseMorph~\cite{PH} trains a diffusion model to manipulate an end-diastolic frame into an end-systolic frame, and interpolate between the latent codes of these two frames to sample intermediate temporal frames during the inference time.
% Also, these I2I models typically rely on interpolation and extrapolation to generate images out of its predefined domain.

% Furthermore, both of these works do not allow their models to be temporally aware of the images at different progression points.

% Recent works: Temporal view 
% Diffusion models
% Linear interploation and/or embeddings of discrete values

%% file: figure/fig_intro.tex
\begin{figure}[t]
    \centering
    \subfloat[\centering Longitudinal Brain MRI]{{\includegraphics[width=0.46\linewidth]{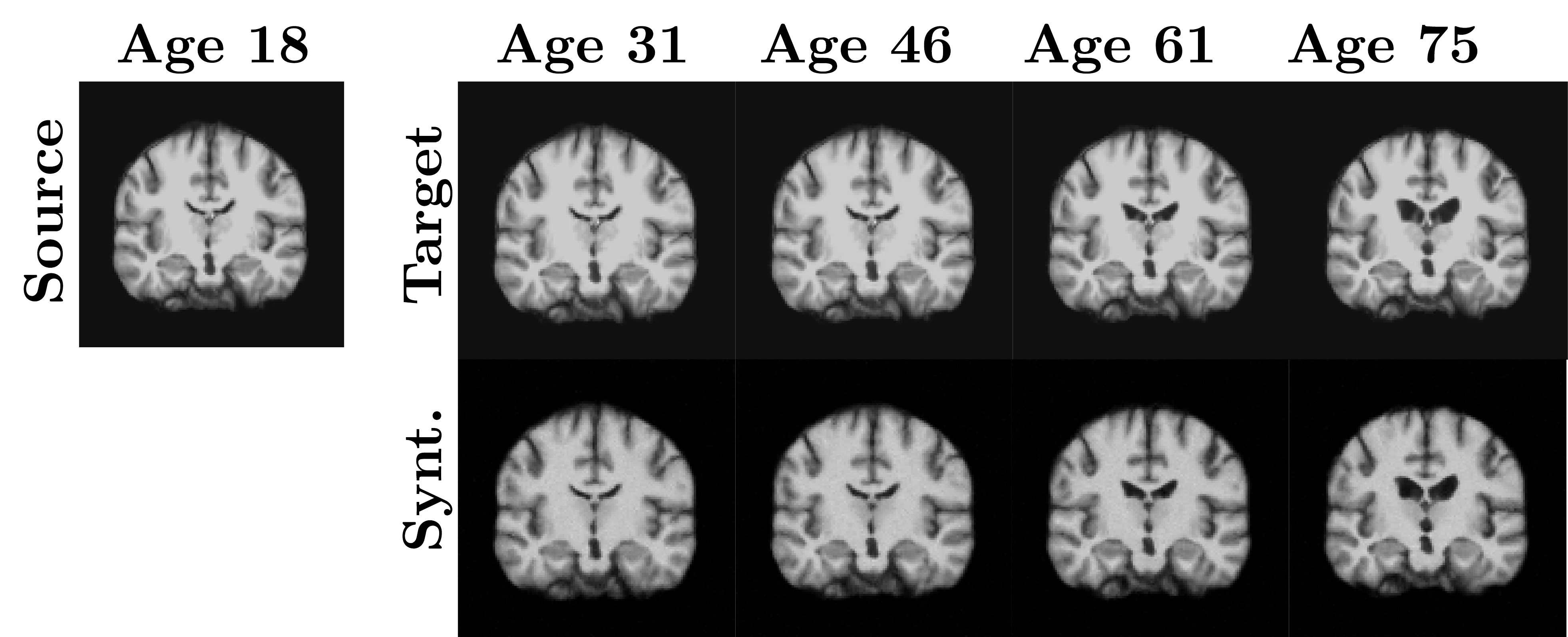} }}
    \qquad
    \subfloat[\centering Multi-frame Cardiac MRI]{{\includegraphics[width=0.46\linewidth]{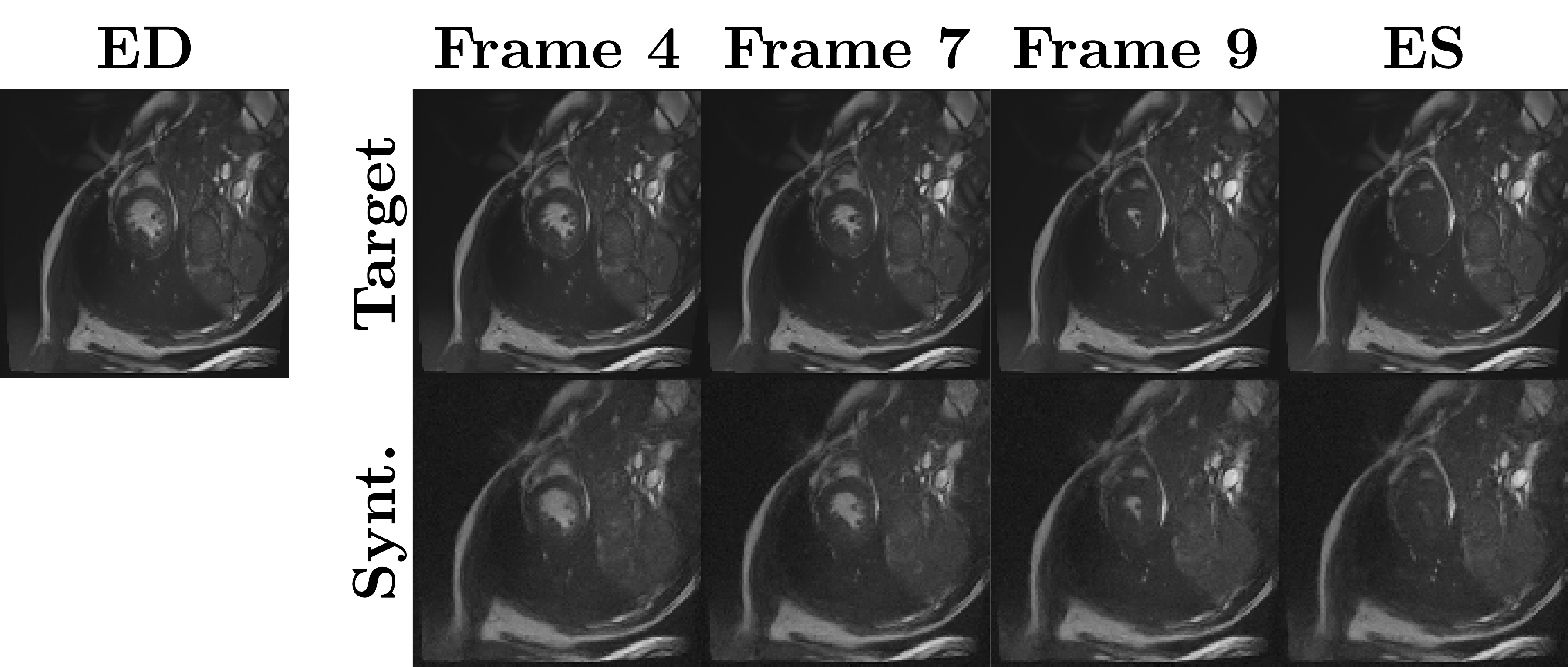} }}
    \caption{
    Examples of longitudinal medical images synthesized by our proposed SADM.
    }
    \label{fig:intro}%
\end{figure}

%% file: section/2_background.tex
\section{Preliminary}\label{sec:background}

\subsection{Diffusion Models}
Diffusion models consist of a forward process, which starts with the data $\x\sim p(\x)$ and gradually adds noise to obtain a noisy version of the data $\z=\left\{\z_t| t\in [0,1]\right\}$, and a reverse process, which reverts the forward process by predicting and subtracting the noise in the reverse direction (\ie, from $t=1$ to $t=0$).

Formally, following~\cite{classifier_free}, we define the forward process $q(\z| \x)$ specified in continuous time $0\leq s<t\leq 1$ as:
\begin{equation}
    q(\z_t| \x)=\mathcal{N}(\alpha_t \x, \sigma_t^2 \mathbf{I}), \quad
    q(\z_t| \z_s)=\mathcal{N}((\alpha_t/\alpha_s)\z_s,\sigma_{t| s}^2\mathbf{I})\label{eq:forward}
\end{equation}
where $\alpha_t^2=1/(1+e^{-t})$ and $\sigma^2_t=1-\alpha_t^2$ are the continuous-time noise schedules, $\sigma^2_{t| s}=\left(1-e^{\lambda_t-\lambda_s}\right)\sigma^2_t$ is the variance term of the $s$ to $t$ transition, and $\lambda_t=\log[\alpha_t^2/\sigma_t^2]$ is the signal-to-noise-ratio of the noise schedules that is monotonically decreasing~\cite{NEURIPS2021_b578f2a5}.
% whose log signal-to-noise-ratio $\lambda_t=\log[\alpha_t^2/\sigma_t^2]$ decreases monotonically.
This forward process can be reformulated in the reverse direction as $q(\z_s|\z_t, \x)=\N(\tilde{\boldsymbol{\mu}}_{s| t}(\z_t,\x), (\tilde{\sigma}^2_{s| t})\mathbf{I})$, where $\tilde{\boldsymbol{\mu}}_{s | t}\left(\z_t, \x\right)=e^{\lambda_t-\lambda_s}\left(\alpha_s / \alpha_t\right) \z_t+\left(1-e^{\lambda_t-\lambda_s}\right) \alpha_s \x$ and $\tilde{\sigma}_{s | t}^2=\left(1-e^{\lambda_t-\lambda_s}\right) \sigma_s^2$.

The reverse process is parameterized by a generative model $\hat{\x}_\theta$ in the form:
\begin{equation}
    p_\theta(\z_s| \z_t)=\N(\tilde{\boldsymbol{\mu}}_{s| t}(\z_t,\hat{\x}_\theta(\z_t, \lambda_t)), \tilde{\mathrm{\Sigma}}_{s|t}\mathbf{I})
\end{equation}
where the variance $\tilde{\mathrm{\Sigma}}_{s|t}=(\tilde{\sigma}^2_{s| t})^{1-v}(\sigma^2_{t| s})^{v}$ is an interpolation between $\tilde{\sigma}^2_{s| t}$ and $\sigma^2_{t| s}$~\cite{NEURIPS2021_b578f2a5}, and $v$ is the hyperparameter that controls the stochasticity of the sampler~\cite{improved_ddpm}.
We use the ancestral sampler~\cite{ddpm} where $\lambda_0<...<\lambda_T=\lambda_1$ for discrete T time steps:
\begin{equation}
    \z_s=\tilde{\boldsymbol{\mu}}_{s \mid t}\left(\z_t, \hat{\x}_\theta\left(\z_t, \lambda_t\right)\right)+\sqrt{\tilde{\mathrm{\Sigma}}_{s|t}} \eps \quad \text{ where } \eps\sim \N(\mathbf{0,I}).\label{eq:sampling}
\end{equation}
% The noise-prediction loss term~\cite{ddpm} for the reverse process generative model is:
% \begin{equation}
%     \mathcal{L}(\x)=\mathbb{E}_{\eps \sim \N(\mathbf{0, I}), t \sim U(0,1)}\left[\left\|\hat{\eps}_\theta\left(\z_t, \lambda_t\right)-\eps\right\|_2^2\right]
% \end{equation}
% where $\z_t=\alpha_t\x+\sigma_t\eps$, and $\hat{\eps}_\theta\left(\z_t, \lambda_t\right)=\sigma^{-1}_t(\z_t-\alpha_t\hat{\x}_\theta(\z_t,\lambda_t))$.

\subsection{Classifier-free Guidance}
There are two branches of conditioning methods for diffusion models: 1) Classifier guided~\cite{NEURIPS2021_49ad23d1}; and 2) Classifier-free guidance~\cite{classifier_free}.
However, it is often hard to define the problem setting for a classifier in the medical domain, and even the state-of-the-art classifiers often do not have the performance suitable for classifier-guided models.
Thus, we opt to use classifier-free guidance for conditioning the diffusion model.
The classifier-free guided diffusion model takes the conditioning signal $\cc$ as an additional input, and is defined as
\begin{equation}
    \tilde{\x}_\theta\left(\z_t, \mathbf{c}, \lambda_t\right)=(1+w) \hat{\x}_\theta\left(\z_t, \mathbf{c}, \lambda_t\right)-w \hat{\x}_\theta\left(\z_t, \boldsymbol{\emptyset}, \lambda_t\right),\label{eq:classifier-free}
\end{equation}
which is the weighted sum of the model with condition $\cc$ and model with zero tensor $\boldsymbol{\emptyset}$, \ie, unconditional model. The guidance strength $w$ controls the trade-off between sample quality and diversity, \ie, the higher the guidance strength the lower the diversity.
Eq.~(\ref{eq:classifier-free}) can also be performed in $\eps$-space $\tilde{\eps}_\theta\left(\z_t, \mathbf{c}, \lambda_t\right)=(1+w) \hat{\eps}_\theta\left(\z_t, \mathbf{c}, \lambda_t\right)-w \hat{\eps}_\theta\left(\z_t, \boldsymbol{\emptyset}, \lambda_t\right)$.

During training, we can randomly replace the conditioning signals $\cc$ by a zero tensor with probability $p_{uncond}$.
Then, the noise-prediction loss term~\cite{ddpm} for the reverse process conditional generative model is:
\begin{equation}\label{eq:loss}
    \mathcal{L}(\x)=\mathbb{E}_{\eps \sim \N(\mathbf{0, I}), t \sim U(0,1), \hat{\mathbf{I}}\sim Be(p_{uncond})}\left[\left\|\hat{\eps}_\theta\left(\z_t, \cc\hat{\mathbf{I}}, \lambda_t\right)-\eps\right\|_2^2\right]
\end{equation}
where $\hat{\mathbf{I}}\sim Be(p_{uncond})$ is either a zero or an identity tensor sampled from a Bernoulli distribution, $\z_t=\alpha_t\x+\sigma_t\eps$, and $\hat{\eps}_\theta\left(\z_t, \cc, \lambda_t\right)=\sigma^{-1}_t(\z_t-\alpha_t\hat{\x}_\theta(\z_t,\cc,\lambda_t))$.

%% file: section/3_method.tex
\section{\ours: Sequence-Aware Diffusion Model}\label{sec:method}
\begin{figure}[t]
\centering
\includegraphics[width=0.9\textwidth]{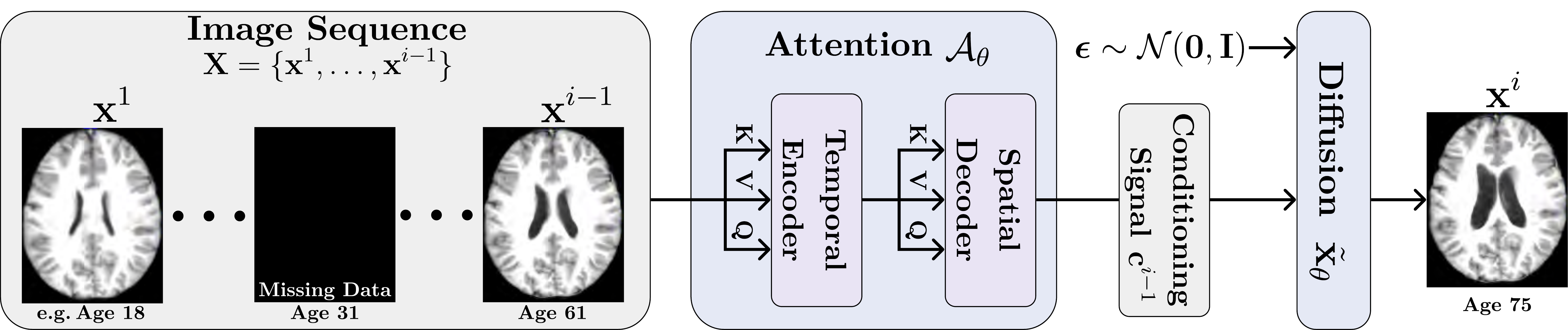}
\caption{An overview of our proposed sequence-aware diffusion model (SADM).} \label{fig:framework}
\end{figure}
We propose a sequence-aware diffusion model (\ours) for longitudinal medical image generation.
Specifically, our proposed SADM uses a transformer-based attention module for conditioning a diffusion model.  
The attention module is a 4D generalization of the video vision transformer (ViViT)~\cite{vivit}, and it is specifically used to generate the conditioning signals for the diffusion model.
In this section, we will briefly explain the problem setting and the details of \ours.
An overview of SADM is illustrated in Fig.~\ref{fig:framework}.

\paragraph{Problem setting}
Let $\xx\sim p(\xx)\in\R^{L\times W\times H\times D}$ be a longitudinal 3D medical image with temporal length $L$.
We partition $\xx$ into conditioning images $\xC\in\R^{n_{\C}\times W\times H\times D}$, missing images $\xM\in\R^{n_{\M}\times W\times H\times D}$, and future images $\xF\in\R^{n_{\F}\times W\times H\times D}$, where $\C=\left\{c_1,\ldots,c_{n_{\C}}\right\}$, $\M=\left\{m_1,\ldots,m_{n_{\M}}\right\}$, and $\F=\left\{f_1,\ldots,f_{n_{\F}}\right\}$ are a sequence of scalar indices for tensor indexing.
We define these sequences as ordered, timestamped, and non-intersecting sets~\cite{Quadrana_2019} such that $\C\cap\M=\C\cap\F=\M\cap\F=\emptyset$, $\C\cup\M\cup\F=\left\{1,...,L\right\}$, and $n_{\C}+n_{\M}+n_{\F}=L$.
We assume that indices of $\F$ are always in future of $\C$ and $\M$, \ie, $c<f$ and $m<f$ for all $c\in\C, m\in\M,$ and $f\in\F$.
Also, we assume that the first image of the sequence is known, \ie, $c_1=1$.
% We only consider cases where we have at lease one conditioning image and one missing or future image, \ie, $n_{\C}\ge1$, $n_{\M}+n_{\F}\ge1$.
The objective is to maximize the posteriors $p(\xM|\xC)$ and $p(\xF|\xC)$, \ie, synthesize the missing and future images given a sequence of conditioning images.

\input{figure/attention_module.tex}
\subsection{Attention Module $\mathcal{A}_\theta$}
Unlike many other longitudinal vision datasets (\eg, video data), longitudinal medical images have the following unique properties: various sequence lengths, missing data or frames, and high dimensionality. existing generative solutions used in common computer vision fields are not optimized for these properties, thus longitudinal medical image generation requires a specialized architecture.
Inspired by the success of transformers for vision datasets and their ability to calculate attention for long-distance spatio-temporal representations~\cite{vivit}, we propose a transformer-based attention module for generating conditioning signals for the diffusion model.
This attention module will direct which frames of the conditioning image $\xC$ will be beneficial to generate the future or missing frames.
An overview of our attention module is illustrated in Fig.~\ref{fig:attention_module}.

\paragraph{Token embedding} A common approach to embedding an image or video into tokens is by running a fusing window over the input with non-overlapping strides~\cite{vivit}.
Given $\xx\in\R^{L\times W\times H\times D}$, we run a non-overlapping linear projection window of dimension $\R^{l\times w\times h\times d\times dim}$ over the image, \ie, a 4D convolution.
The resulting unflattened token $\h$ have the shape of $\R^{\lfloor\frac{L}{l}\rfloor\times\lfloor\frac{W}{w}\rfloor\times\lfloor\frac{H}{h}\rfloor\times\lfloor\frac{D}{d}\rfloor\times dim}$.
However, as we are dealing with longitudinal medical images that may contain missing frames, fusing through the temporal axis is not feasible.
Thus, we set $l=1$ for experiments conducted in this paper.
Also, the temporal resolution of longitudinal images is typically very low compared to its spatial resolution, so setting $l=1$ will only slightly affect the computational efficiency.

\paragraph{Temporal encoder}
Inspired by factorized transformers~\cite{vivit}, we factorize our attention module into a temporal encoder and a spatial decoder that can benefit from long-range spatio-temporal attention with high computational efficiency.
The temporal encoder computes the self-attention temporally among all tokens in the same spatial index.
Specifically, it takes the unflattened token and reshapes them into $\h_\text{tmp}\in\R^{\lfloor\frac{W}{w}\rfloor\cdot\lfloor\frac{H}{h}\rfloor\cdot\lfloor\frac{D}{d}\rfloor\times L\times dim}$, where the leading dimension is the batch dimension.
Then, it computes the self-attention along the temporal dimensions, and an MLP reduces the token's dimension by a factor of $2^3$ and reshapes them to $\h^{\ell}_\text{tmp}\in\R^{\lfloor\frac{W}{2^\ell w}\rfloor\cdot\lfloor\frac{H}{2^\ell h}\rfloor\cdot\lfloor\frac{D}{2^\ell d}\rfloor\times L\times dim}$ for each temporal transformer block $\ell$.
% Finally, we perform global average pooling operation on the spatial dimensions to obtain the reshaped temporal token $\h_\text{tmp}\in\R^{1\times L\times dim}$.

\input{others/alg_training.tex}
\paragraph{Spatial decoder}
The output of the temporal encoder is reshaped into a spatial token $\h_{\text{spt}}\in\R^{L\times\lfloor\frac{W}{2^N w}\rfloor\cdot\lfloor\frac{H}{2^N h}\rfloor\cdot\lfloor\frac{D}{2^N d}\rfloor\times dim}$, and the spatial decoder calculates self-attention to spatial dimensions between all tokens in the same temporal index.
Then, it is upsampled by a factor of $2$ for each spatial dimension to obtain $\h^{\ell}_{\text{spt}}\in\R^{L\times\lfloor\frac{W}{2^{N-\ell} w}\rfloor\cdot\lfloor\frac{H}{2^{N-\ell} h}\rfloor\cdot\lfloor\frac{D}{2^{N-\ell} d}\rfloor \times dim}$.
Finally, we unflatten and reshape the output of the last block into $\R^{\lfloor\frac{W}{w}\rfloor\times\lfloor\frac{H}{h}\rfloor\times\lfloor\frac{D}{d}\rfloor\times L\cdot dim}$, and perform upsampling and a 3D convolution operation to obtain the conditioning signal $\cc\in\R^{W\times H\times D}$. 

Since transformers can mask specific indexes of a token, we can train and infer even when there are missing frames in the longitudinal images.
However, we have found that using zero tensors for missing frames with non-zero positional encoding performs better than masking missing frames.

\subsection{Conditional Diffusion Model}
Our proposed SADM follows the formulation of classifier-free diffusion model~\cite{classifier_free} defined in Section~\ref{sec:background}.
We extend this diffusion model by using a sequence-aware conditioning signal explained in the previous section.
Furthermore, we use an autoregressive sampling scheme that can effectively capture the long-distance temporal dependency during inference.

\paragraph{Training SADM}
During training, the input to the diffusion model is a randomly selected target image from unobserved indices $\M\cup\F$ and a conditioning signal from previous indices, \ie, $\x^i\in\R^{W\times H\times D}$ and $\cc=\mathcal{A}(\{\x^1,...,\x^{i-1}\})$, respectively.
The attention module and the conditional diffusion model can be pretrained separately and finetuned together or trained end-to-end from scratch with the loss term defined in Eq.~\ref{eq:loss}.
The attention module can be pretrained by minimizing the $\ell_2$ loss between the target image and the conditioning signal $\cc$, and the diffusion model can be pretrained with a zero or random-valued tensor as a conditioning signal.
However, we have found that training end-to-end from scratch performs better.
Our training pipeline is defined in Algorithm~\ref{alg:training}.

\paragraph{Autoregressive sampling}
\input{others/alg_sampling.tex}
Our SADM samples the next-frame image $\x^i$ given the conditional signals of its previous images, \ie, $\cc^{i-1}=\mathcal{A}_\theta(\{\x^1,...,\x^{i-1}\})$.
However, real-world data often have missing data or only a single image per subject.
Thus, we use an autoregressive sampling scheme that imputes missing images with synthesized images autoregressively.
This autoregressive sampling scheme has been shown to improve the generative performance of diffusion models~\cite{harvey2022flexible}.
An overview of the autoregressive sampling scheme is shown in Algorithm~\ref{alg:sampling}.

%% file: figure/attention_module.tex
\begin{figure}[t]
\begin{center}
\includegraphics[width=1\textwidth]{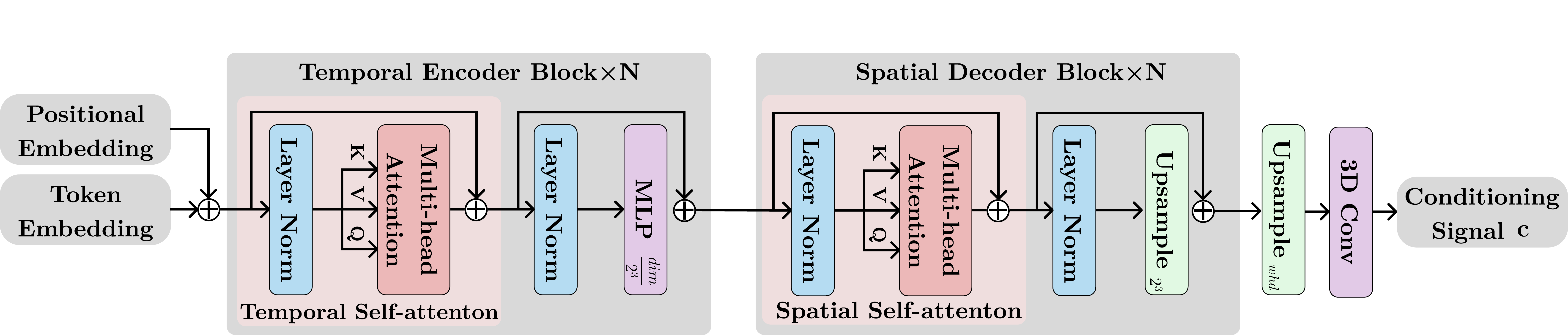}
\end{center}
\caption{
An illustration of the attention module $\mathcal{A}_\theta$.
The temporal encoder performs temporal self-attention followed by dimension reduction using an MLP, while the spatial decoder performs a spatial self-attention followed by an upsampling operation.
} \label{fig:attention_module}
\end{figure}

%% file: others/alg_training.tex
\begin{algorithm}[t]
\caption{SADM Training}\label{alg:training}
\Repeat{\upshape converged}{
Initialize an empty sequence of images $\tilde{\xx}$\\
$(\xx,\C,\M,\F)\sim p(\xx)$ \tcp{Sample data and indices}
$i\sim U(\M\cup\F)$ \tcp{Randomly select target index}
\For{$k=1,...,i-1$}{
    \If{$k\in\C$}{
        $\tilde{\xx}\leftarrow \tilde{\xx}+[deque(\xC)]$ \tcp{Append conditioning image}
    }
    \Else{
        $\tilde{\xx}\leftarrow \tilde{\xx}+[\boldsymbol{\emptyset}]$ \tcp{Append zero tensor}
    }
}
$\cc^{i-1}\leftarrow \mathcal{A}_{\theta}(\tilde{\xx})$ \tcp{Generate conditioning signal}
% \tcp{$\tilde{\xx}\in \R^{(i-1)\times W\times H\times D}, \cc^{i-1}\in\R^{W\times H\times D}$}
$(\boldsymbol{\epsilon}, t, \hat{\mathbf{I}})\sim (\N(\mathbf{0,I}), U(0,1), Be(p_{uncond}))$\\
$\z_t\leftarrow\alpha_t\xx^i+\sigma_t\boldsymbol{\epsilon}$ \tcp{Sample noisy image}
% $\cc^{i-1}\leftarrow \boldsymbol{\emptyset}$ with probability $p_{uncond}$ \tcp{For classifier-free guidance}
Take gradient step on $\nabla_{\theta}\|\hat{\boldsymbol{\epsilon}}_\theta\left(\z_t, \cc^{i-1}\hat{\mathbf{I}}, \lambda_t\right)-\boldsymbol{\epsilon}\|_2^2$
}
\end{algorithm}

%% file: others/alg_sampling.tex
\begin{algorithm}[t]
\caption{Autoregressive Sampling}\label{alg:sampling}
\textbf{Input:} Conditioning indices $\C=\{c_1,...,c_{n_\C}\}$ and images $\xC$\\
\textbf{Output:} Missing images $\xM$ and future images $\xF$\\
Initialize an empty sequence of images $\tilde{\xx}$\\
 \For{$i=1,...,L$}{
    \If(\tcp*[h]{We assume $c_1=1$}){$i \in \C$}{
      $\tilde{\xx}\leftarrow \tilde{\xx}+[deque(\xC)]$
    }
    \Else{
      $\cc^{i-1}\leftarrow \mathcal{A}_{\theta}(\tilde{\xx})$ \tcp{$\tilde{\xx}=[\x^1,...,\x^{i-1}]$}
      $\z_1\leftarrow \N(\mathbf{0,I})$\\
      \For{$t=1,...,\frac{2}{T},\frac{1}{T}$}{ 
      $s\leftarrow t-\frac{1}{T}$\\
      $\boldsymbol{\epsilon}\sim \N(\mathbf{0,I})$\\
      $\z_s\leftarrow\tilde{\boldsymbol{\mu}}_{s \mid t}\left(\z_t, \tilde{\x}_\theta\left(\z_t, \mathbf{c}^{i-1}, \lambda_t\right)\right)+\sqrt{\tilde{\mathrm{\Sigma}}_{s|t}} \boldsymbol{\epsilon}$ \tcp{Eq. (\ref{eq:sampling}) and (\ref{eq:classifier-free})}
      }
      $\tilde{\x}^{i}\leftarrow \z_0$\\ 
      $\tilde{\xx}\leftarrow \tilde{\xx}+[\tilde{\x}^{i}]$ \tcp{$\tilde{\xx}=[\x^1,...,\x^{i-1}]+[\tilde{\x}^{i}]$}
    }
 }
$\xC,\xM,\xF\leftarrow partition(\tilde{\xx})$\\
\textbf{return} $\xM,\xF$
% \textbf{return} $\xx$ \tcp{Or partition into and return $\xC, \xM, \xF$}
\end{algorithm}

%% file: section/4_experiment.tex
\section{Experiments}
In this section, we show the effectiveness of our proposed SADM in medical image generation on one public 3D longitudinal 
 cardiac MRI dataset and one simulated 3D longitudinal 
 brain MRI dataset.
% First, we show a proof of concept using two toy datasets: 1) Rotating RaFD face dataset~\cite{rafd}; and 2) Synthetic longitudinal brain MRI.
% Then, 
We compare our work with GAN-based~\cite{gan-based} and diffusion-based~\cite{diffusemorph} baselines quantitatively and qualitatively.
Finally, an ablation study of our model components with various settings for the input sequence is presented.

\input{figure/fig_baseline.tex}
\subsection{Dataset and Implementation}\label{sec:datset}
% \paragraph{ACDC cardiac MRI}

\noindent\textbf{Cardiac dataset.} We use the multi-frame cardiac MRI curated by ACDC (Automated Cardiac Diagnosis Challenge) organizers~\cite{acdc}.
A common task for cardiac image generation is to synthesize the final frame of a cardiac cycle (\ie, end-systolic or ES), given a starting frame of the cycle (\ie, end-diastolic or ED).
% However, ACDC dataset consists of 4D (3D+t) cardiac MRI with multiple frames in between ED (end-diastolic) frame and ES (end-systolic) frame.
The ACDC dataset consists of cardiac MRI from 100 training subjects and 50 testing subjects.
We take intermediate frames from ED to ES and resize them to $\xx\in\R^{12\times128\times128\times32}$, where each dimension is the length of the frame, the width, the height, and the depth, respectively.
Then we Min-Max normalize the dataset subject-wise.
Although MRI resizing results in uneven resolution, we opt for this approach, as it is the most reproducible preprocessing method.
For training, we randomly select conditioning, missing, and future indices.
During inference, we experiment with three settings: 1) Single image, where only the ED frame is given as input; 2) Missing data, where the input sequence has randomly missing frames; and 3) Full sequence, where the input sequence is fully loaded with conditioning images.

\noindent\textbf{Brain dataset.} Simulating healthy subjects' brain changes over time is essential for understanding human aging~\cite{feng2020estimating}. The in-house synthesis of longitudinal brain MRI was carried out in two main steps, using 2,851 subject scans evenly distributed in age.
% \footnote{A separate article for this dataset is being prepared at the time of writing this paper, so we briefly explain the dataset and use if only for proof of concept.}.
We first divided these subject scans into five age groups (18-30, 31-45, 46-60, 61-74, and 75-97 years old) and generated five age-specific templates following~\cite{Zhang_2023}.
Then we used a GAN-based registration model to register each subject scan to these five templates, respectively, to simulate the longitudinal images of the same person at different ages~\cite{Zhang_2023b}.
Templates and registered images were divided into ten cross-validation folds.

\noindent\textbf{Implementation.}
We follow the classifier-free diffusion model~\cite{classifier_free} architecture and hyperparameters, and modify the model into a 3D model.
For the transformer, we use the spatial and temporal transformer blocks introduced in ViVit~\cite{vivit} (specifically, Model 3 of ViVit).
We trained the model for 3 million iterations, which took about 150 GPU hours using Nvidia V100 32GB GPUs.
For inference, we use diffusion time steps of $T=1,000$ and classifier-free guidance of $w=0.1$.
\iffalse
The source codes for SADM can be found in \url{https://anonymized.for.review.com}.
\fi

\subsection{Comparison with Baseline Methods}
\input{others/table_baseline.tex}
We choose two state-of-the-art baseline models for comparison: 1) GAN-based model~\cite{gan-based}, which uses a UNet-based GAN model to synthesize an ES frame given an ED frame; and 2) Diffusion-based model~\cite{diffusemorph}, which uses a diffusion model with a deep registration model to register an ED frame into an ES frame.
We follow the same training and inference pipeline for cardiac and brain image generation, so we will explain the settings only for a cardiac dataset as follows.
For training, SADM uses the intermediate frames between ED and ES, so we augment the baselines with pairs of intermediate frames and its ES frame for a fair comparison.
Since these models can only perform single image synthesis (\ie, ED to ES translation), we follow their inference pipeline using only the ED frame as input.
We used the source code provided in each respective paper, only modifying the data loader for preprocessing and augmentation.
% Hence our re-implemented results are slightly different from the results in their published paper.

A qualitative comparison is presented in Fig.~\ref{fig:baseline}, and a quantitative comparison is presented in Table~\ref{tab:baseline}.
For cardiac image generation, our proposed SADM shows a better depiction of the blood pool (red box) compared to the baseline methods.
Also, other areas surrounding the blood pool, such as the myocardium and ventricles, are also synthesized with higher fidelity.
For brain image generation, the ventricular regions (blue box) synthesized by SADM are more crisp compared to baselines, and the cortical surface is synthesized more accurately. 
Then, we perform a quantitative comparison by calculating the structural similarity index (SSIM), peak signal-to-noise ratio (PSNR), and normalized root-mean-square deviation (NRMSE) between the target and the synthesized ES frame.
Our model outperforms the GAN-based method~\cite{gan-based} by $3\sim13\%$ in each metric while slightly outperforming the diffusion-based model~\cite{diffusemorph}.
It is worth noting that the diffusion-based baseline uses a source image and a reference image for registration, whereas we only use the source image as input.
Although our proposed SADM is capable of working with a single image, it is designed to perform even better with a sequence of images, as shown in the next section.
% In our defense, the diffusion-based baseline uses a source image and a reference image for registration, while we only use the source image as input.
% Although our proposed SADM can certainly work with a single image, it is designed to perform better with a sequence of images, as presented in the next section.

\subsection{Ablation Study}
\input{figure/fig_ablation.tex}
In this section, we perform an ablation study on the components of our model with various settings for the input sequences.
First, we experiment with different settings for the input sequence defined in the first paragraph of Section~\ref{sec:datset}, \ie, single image, missing data, and full sequence settings.
As presented in Fig.~\ref{fig:ablation}, synthesis using the full sequence and missing data settings show a higher SSIM compared to a single input setting.
Also, as observed by the high peak in SSIM for frames in the vicinity of conditioning frames, our SADM is learning which frames of the input sequence are important in generating future frames, \ie, the sequential dependency.
Next, we perform an ablation study by removing either the attention module or the diffusion model.
\input{others/table_ablation.tex}
The diffusion-only model can be trained with the raw pixel values of the sequential image as conditioning signals, and the attention-only model can be trained by minimizing the $\ell_2$ loss between the target image and the output of the transformer.
As shown in Table~\ref{tab:ablation}, evidently, the attention-only module has the worst performance as the transformers are not designed for image generation (typically due to flattening operations~\cite{Zhang_2021}).
The diffusion-only model performs on par with GAN-based baseline~\cite{gan-based}, but it is unable to learn the sequential dependency, as observed by the minimal performance increase in full sequence setting compared to single input settings.

%% file: figure/fig_baseline.tex
\begin{figure}[t]
\begin{center}
\includegraphics[width=1\textwidth]{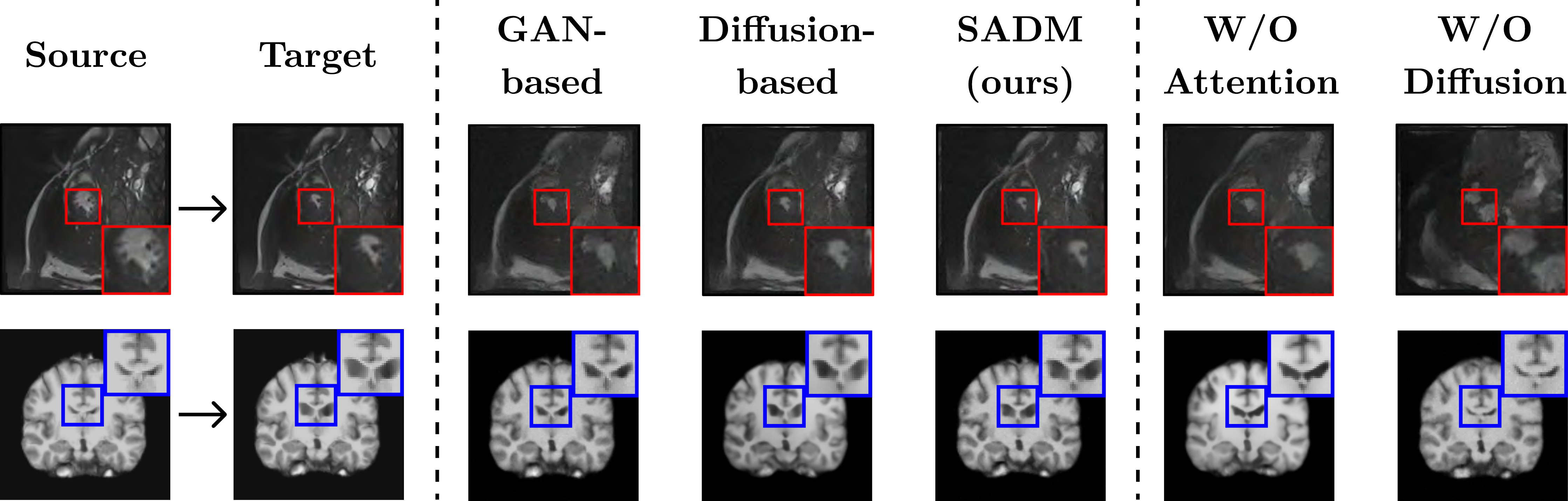}
\end{center}
\caption{
Qualitative comparison between baselines, proposed SADM, and ablated models for single image setting.
GAN-based model~\cite{gan-based} uses a UNet-based GAN model to synthesize MRI frame, and diffusion-based model~\cite{diffusemorph} uses a diffusion model with a deep registration model to register one frame to another.
W/O Attention model is SADM with only the diffusion model, while W/O diffusion model is SADM with only the attention module.
Red boxes indicate blood pool regions and blue boxes indicate ventricle regions.
} \label{fig:baseline}
\end{figure}

%% file: others/table_baseline.tex
% Please add the following required packages to your document preamble:
% \usepackage{booktabs}
% \usepackage{multirow}
\begin{table}[t]
\centering
\caption{Quantitative comparison between baselines and our propose SADM.}
\label{tab:baseline}
\begin{tabular}{@{}ccccccc@{}}
\toprule
\multirow{2}{*}{Method}             & \multicolumn{3}{c}{Cardiac MRI} & \multicolumn{3}{c}{Brain MRI} \\ \cmidrule(l){2-7} 
     & SSIM $\uparrow$ & PSNR $\uparrow$ & NRMSE $\downarrow$ & SSIM $\uparrow$ & PSNR $\uparrow$ & NRMSE $\downarrow$ \\ \midrule
GAN-based \cite{gan-based}          & 0.788   & 27.394   & 0.176  &     0.955    &    30.502    &   0.138     \\
Diffusion-based \cite{diffusemorph} & 0.842   & 28.863   & 0.154  &   0.961      &  31.229      &    0.121    \\
Ours & \textbf{0.851}  & \textbf{28.992} & \textbf{0.153}     & \textbf{0.978}       & \textbf{31.699}       & \textbf{0.090}          \\ \bottomrule
\end{tabular}
\end{table}

%% file: figure/fig_ablation.tex
\begin{figure}[t]
\begin{center}
\includegraphics[width=1\textwidth]{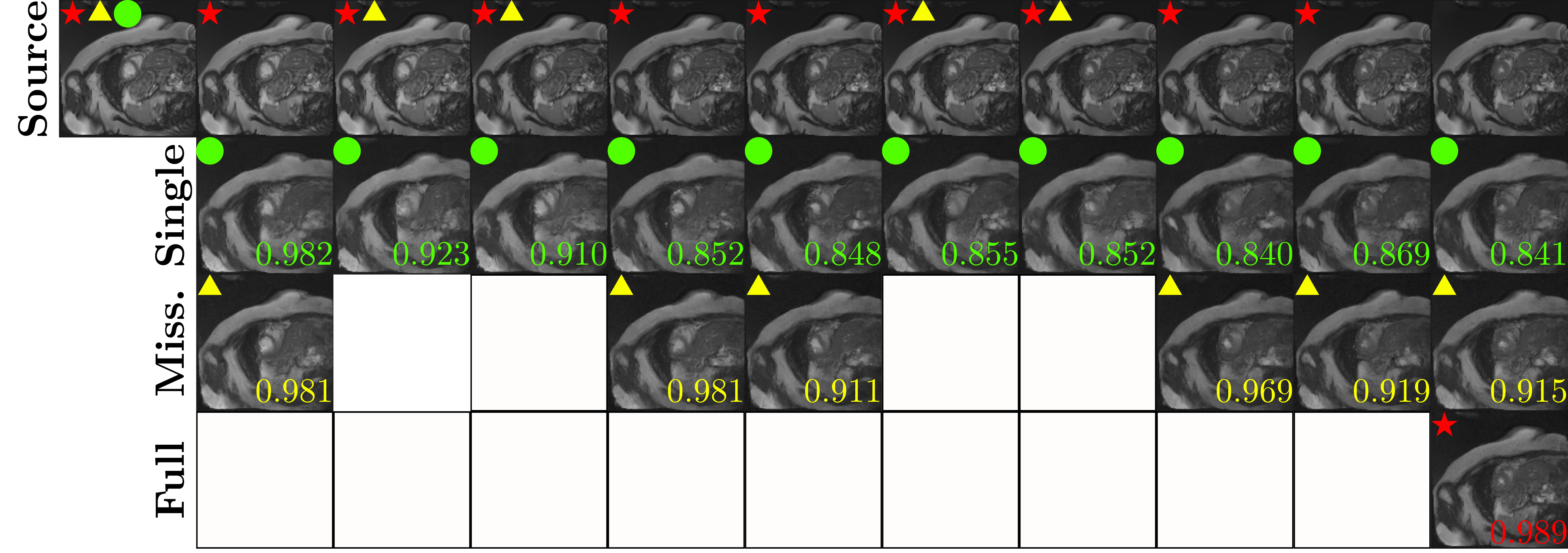}
\end{center}
\caption{
An illustration of different settings for input sequence. The first row shows the ground-truth progression from ED frame to ES frame.
The symbols in upper left corner of images in the first row is the conditioning image $\xC$ (red star for full sequence, yellow triangle for missing data, and green circle for single image).
The remaining rows show the synthesized images with single image, missing data, and full sequence settings, respectively.
The numbers on bottom right corner of each image is the SSIM between the ground truth and the synthesized frame.
} \label{fig:ablation}
\end{figure}

%% file: others/table_ablation.tex
\begin{table}[t]
\centering
\caption{An ablation study of SADM components with single image, missing data, and full sequence settings using the ACDC cardiac dataset.}
\label{tab:ablation}
\begin{tabular}{@{}cc|ccc|ccc|ccc@{}}
\toprule
\multicolumn{2}{c|}{Method} & \multicolumn{3}{c|}{SSIM $\uparrow$} & \multicolumn{3}{c|}{PSNR $\uparrow$} & \multicolumn{3}{c}{NRMSE $\downarrow$} \\ \midrule
\multicolumn{1}{l}{Diffusion} & Attention & Single      & Missing     & Full        & Single      & Missing     & Full        & Single      & Missing     & Full        \\ \midrule
\checkmark & \checkmark & 0.851  & 0.916  & 0.977  & 28.992  & 30.314 & 33.733  & 0.153   & 0.143 & 0.087  \\
\checkmark &           & 0.778 & 0.792 & 0.802 & 26.955 & 26.223  & 28.031  & 0.179 & 0.175 & 0.167 \\
          & \checkmark & 0.703 & 0.726 & 0.727 & 25.533 & 25.629 & 25.275 & 0.267 & 0.234 & 0.217 \\ \bottomrule
\end{tabular}
\end{table}

%% file: section/5_conclusion.tex
\section{Conclusion}
% \paragraph{Limitations}
% Diffusion models are computationally hungry models, as they need to iteratively denoise multiple times to sample an image.
% Our proposed SADM uses an autoregressive sampling scheme, which needs to synthesize images in the previous indexes to generate the target image.
% Alternative self-conditioning sampling methods for the video diffusion model have been explored.% in~\cite{harvey2022flexible}.
% However, most of these are suitable for high temporal resolution data, which is typically not the case for the medical domain.
% Sampling efficiency of diffusion models is a very active research topic, so we hope to find a solution for this in the near future.

To this end, we propose a sequence-aware diffusion model for the generation of longitudinal medical images.
Specifically, our model consists of a transformer-based attention module that can learn the sequential or temporal dependence of longitudinal data input and a diffusion model that can synthesize high-fidelity medical images.
We tested our proposed SADM on longitudinal cardiac and brain MRI generation and presented state-of-the-art performance quantitatively and qualitatively.
Our approach to learning the temporal dependence of sequential data and using it as a prior in diffusion models is an exciting new research topic in the field of medical image generation.
However, the limitations of our model's computational efficiency for large medical datasets suggest that further work is needed to improve sampling efficiency.
We hope our research inspires researchers to pursue this newly found topic and find solutions to these challenges.
% To the best of our knowledge, we are one of the first to explore the temporal dependency of sequential data and use it as a prior in diffusion models for medical image generation.
% Our SADM model achieves state-of-the-art longitudinal medical image generation, but future work may aim to improve computational efficiency, particularly for large medical datasets, making this an exciting new research topic.
% We hope to inspire researchers to actively engage in this newly found research topic.

%% file: section/6_acknowledgement.tex
\subsubsection{Acknowledgments}
This work is supported in part by the Natural Sciences and Engineering Research Council of Canada (NSERC), and NVIDIA Hardware Award, and Institute of Information \& communications Technology Planning \& Evaluation (IITP) grant funded by the Korea government (MSIT) No. 2022-0-00959 ((Part 2) Few-Shot Learning of Causal Inference in Vision and Language for Decision Making), and the MOTIE (Ministry of Trade, Industry, and Energy) in Korea, under Human Resource Development Program for Industrial Innovation (Global) (P0017311) supervised by the Korea Institute for Advancement of Technology (KIAT).

%% file: main.bbl
\begin{thebibliography}{10}
\providecommand{\url}[1]{\texttt{#1}}
\providecommand{\urlprefix}{URL }
\providecommand{\doi}[1]{https://doi.org/#1}

\bibitem{vivit}
Arnab, A., et~al.: Vivit: A video vision transformer (2021)

\bibitem{8633930}
Balakrishnan, G., et~al.: Voxelmorph: A learning framework for deformable
  medical image registration. IEEE Transactions on Medical Imaging
  \textbf{38}(8),  1788--1800 (2019)

\bibitem{acdc}
Bernard, O., et~al.: Deep learning techniques for automatic {MRI} cardiac
  multi-structures segmentation and diagnosis: Is the problem solved? {IEEE}
  Transactions on Medical Imaging  \textbf{37}(11),  2514--2525 (nov 2018)

\bibitem{gan-based}
Campello, V.M., et~al.: Cardiac aging synthesis from cross-sectional data with
  conditional generative adversarial networks. Frontiers in Cardiovascular
  Medicine  \textbf{9} (sep 2022)

\bibitem{NEURIPS2021_49ad23d1}
Dhariwal, P., Nichol, A.: Diffusion models beat gans on image synthesis. In:
  Advances in Neural Information Processing Systems. vol.~34, pp. 8780--8794
  (2021)

\bibitem{feng2020estimating}
Feng, X., et~al.: Estimating brain age based on a uniform healthy population
  with deep learning and structural magnetic resonance imaging. Neurobiology of
  aging  \textbf{91},  15--25 (2020)

\bibitem{harvey2022flexible}
Harvey, W., et~al.: Flexible diffusion modeling of long videos. In: Advances in
  Neural Information Processing Systems (2022)

\bibitem{ddpm}
Ho, J., Jain, A., Abbeel, P.: Denoising diffusion probabilistic models. In:
  Advances in Neural Information Processing Systems. vol.~33, pp. 6840--6851
  (2020)

\bibitem{classifier_free}
Ho, J., Salimans, T.: Classifier-free diffusion guidance (2022)

\bibitem{imagen_video}
Ho, J., et~al.: Imagen video: High definition video generation with diffusion
  models (2022)

\bibitem{causal_ml}
Kaddour, J., et~al.: Causal machine learning: A survey and open problems (2022)

\bibitem{diffusemorph}
Kim, B., Han, I., Ye, J.C.: {DiffuseMorph}: Unsupervised deformable image
  registration using diffusion model. In: ECCV. pp. 347--364 (2022)

\bibitem{kim2022diffusion}
Kim, B., Ye, J.C.: Diffusion deformable model for 4d temporal medical image
  generation. In: Medical Image Computing and Computer Assisted
  Intervention--MICCAI 2022: 25th International Conference, Singapore,
  September 18--22, 2022, Proceedings, Part I. pp. 539--548. Springer (2022)

\bibitem{NEURIPS2021_b578f2a5}
Kingma, D., et~al.: Variational diffusion models. In: Advances in Neural
  Information Processing Systems. vol.~34, pp. 21696--21707 (2021)

\bibitem{improved_ddpm}
Nichol, A.Q., Dhariwal, P.: Improved denoising diffusion probabilistic models.
  In: ICML. vol.~139, pp. 8162--8171 (18--24 Jul 2021)

\bibitem{9854196}
Oh, K., Yoon, J.S., Suk, H.I.: Learn-explain-reinforce: Counterfactual
  reasoning and its guidance to reinforce an alzheimer's disease diagnosis
  model. IEEE Transactions on Pattern Analysis and Machine Intelligence pp.
  1--15 (2022)

\bibitem{pinaya_2022}
Pinaya, W.H.L., et~al.: Brain imaging generation with latent diffusion models.
  In: Deep Generative Models (2022)

\bibitem{Quadrana_2019}
Quadrana, M., Cremonesi, P., Jannach, D.: Sequence-aware recommender systems.
  {ACM} Computing Surveys  \textbf{51}(4),  1--36 (jul 2019)

\bibitem{Yi_2019}
Yi, X., Walia, E., Babyn, P.: Generative adversarial network in medical
  imaging: A review. Medical Image Analysis  \textbf{58},  101552 (dec 2019)

\bibitem{Zhang_2023}
Zhang, C., et~al.: Constructing age-specific mri brain templates based on a
  uniform healthy population across life span with transformer. In: 2023 ISMRM
  and SMRT Annual Meeting \& Exhibition (2023)

\bibitem{Zhang_2023b}
Zhang, C., et~al.: Cycle inverse consistent deformable medical image
  registration with transformer. In: ISMRM and SMRT Annual Meeting \&
  Exhibition (2023)

\bibitem{Zhang_2021}
Zhang, X., et~al.: {RSTNet}: Captioning with adaptive attention on visual and
  non-visual words. In: CVPR. {IEEE} (jun 2021)

\end{thebibliography}
